\documentclass[letterpaper]{article} 
\usepackage[draft]{aaai2026}
\usepackage{times}  
\usepackage{helvet}  
\usepackage{courier}  
\usepackage[hyphens]{url}  
\usepackage{graphicx} 
\usepackage{natbib}  
\usepackage{caption} 
\usepackage{algorithm}
\usepackage{algorithmic}

\usepackage{amsmath,amsfonts,bm}

\def\eqref#1{equation~\ref{#1}}

\def\1{\bm{1}}

\def\vx{{\bm{x}}}
\def\vy{{\bm{y}}}

\DeclareMathAlphabet{\mathsfit}{\encodingdefault}{\sfdefault}{m}{sl}
\SetMathAlphabet{\mathsfit}{bold}{\encodingdefault}{\sfdefault}{bx}{n}

\def\gN{{\mathcal{N}}}

\usepackage{makecell}
\usepackage{multirow}
\usepackage{pifont}
\usepackage{colortbl}
\usepackage{booktabs} 
\usepackage{newfloat}
\usepackage{listings}
\DeclareCaptionStyle{ruled}{labelfont=normalfont,labelsep=colon,strut=off} 
\floatstyle{ruled}
\newfloat{listing}{tb}{lst}{}
\floatname{listing}{Listing}
\title{Bi-level Personalization for Federated Foundation Models: \\ A Task-vector Aggregation Approach}
\author{
     Yiyuan Yang\textsuperscript{\rm 1}, Guodong Long\textsuperscript{\rm 1}, Qinghua Lu\textsuperscript{\rm 2}, Liming Zhu\textsuperscript{\rm 2}, Jing Jiang\textsuperscript{\rm 1}
}
\affiliations{

    \textsuperscript{\rm 1}University of Technology Sydney, Australia\\
    \textsuperscript{\rm 2}CSIRO’s Data61, Australia\\
    Yiyuan.Yang-1@student.uts.edu.au, \{guodong.long, jing.jiang\}@uts.edu.au,\\ \{Qinghua.Lu, Liming.Zhu\}@data61.csiro.au
}

\usepackage{bibentry}

\begin{document}

\maketitle

\begin{abstract}
Federated foundation models represent a new paradigm to jointly fine-tune pre-trained foundation models across clients. It is still a challenge to fine-tune foundation models for a small group of new users or specialized scenarios, which typically involve limited data compared to the large-scale data used in pre-training. In this context, the trade-off between personalization and federation becomes more sensitive. To tackle these, we proposed a bi-level personalization framework for federated fine-tuning on foundation models. Specifically, we conduct personalized fine-tuning on the client-level using its private data, and then conduct a personalized aggregation on the server-level using similar users measured by client-specific task vectors. Given the personalization information gained from client-level fine-tuning, the server-level personalized aggregation can gain group-wise personalization information while mitigating the disturbance of irrelevant or interest-conflict clients with non-IID data. The effectiveness of the proposed algorithm has been demonstrated by extensive experimental analysis in benchmark datasets.
\end{abstract}


\section{Introduction}
Considering the exhaustion of publicly available data and the growing importance of data privacy, recent studies have begun to explore the adaptation of foundation models within the federated learning (FL) framework to leverage decentralized private data for collaborative learning \cite{zhuang2023foundation}. This emerging direction, known as Federated Foundation Models (FedFM), primarily focuses on fine-tuning pre-trained foundation models in small-scale federated settings, in contrast to the massive scale of pre-training. Typically, FedFM fine-tuning involves a limited number of new users or specialized scenarios, aiming to align the general knowledge of the pre-trained model with specific local contexts. Thus, the emphasis in FedFM fine-tuning shifts toward personalization and local adaptation, posing unique challenges in designing effective and efficient methods.

To enable effective personalization in FedFM, various methods have been proposed, including the incorporation of additional personalized modules to align client-specific distribution for personalization \cite{chen2024feddat,yang2024dual} and methods that explicitly decouple global and personalized components to facilitate more flexible adaptation \cite{guo2024selective}.
While effective, these methods primarily focus on client-level personalization, relying on uniform or heuristically fixed federated aggregation at server-side for global model updates. However, in highly heterogeneous FedFM settings, such conventional federation may undermine personalization performance. Specifically, aggregating irrelevant or interest-conflict tasks among clients may dilute task-specific knowledge, while neglecting to amplify contributions from similar tasks in clients may hinder the benefit from effective knowledge sharing for group-wise personalization gaining. Furthermore, many existing approaches introduce additional personalized modules in clients, which often incur non-trivial computational and storage overhead. This presents a significant challenge in FedFM, where foundation models are typically large-scale and clients may have limited resources. Therefore, it is vital to carefully consider the trade-off between personalization and federation for efficient and effective personalization in FedFM.

Although several aggregation-weighting strategies have been proposed in conventional FL to balance personalization and federation by adjusting aggregation weights based on client heterogeneity or similarity inferred from local model parameters \cite{huang2021personalized,zhang2023fedala,rehman2023dawa}, they may be suboptimal for FedFM due to the limited parameter variation across clients, a unique characteristic of foundation model fine-tuning. Foundation models are typically pre-trained on large-scale, diverse datasets, providing a strong and generalizable initialization for various downstream tasks fine-tuning. As a result, local fine-tuning in FedFM usually involves small parameter updates, leading to highly similar model parameters across clients. This significantly limits the effectiveness of conventional FL methods that rely on parameter divergence to infer client similarity or discrepancy. Consequently, there is a pressing need to develop new methods tailored to FedFM, which can effectively balance personalization and federation under high data heterogeneity and constrained parameter variance.

To fill this gap, we propose a bi-level personalization framework for FedFM fine-tuning, named FedBip, which jointly leverages both client-level personalization and server-level personalization to enhance model adaptation in heterogeneous settings. At the client level, FedBip fine-tunes the foundation model on each client's local dataset, enabling task-specific adaptation and personalization. At the server level, to further improve personalization, we introduce a task-vector aggregation mechanism to promote collaboration among similar tasks and mitigate interference from irrelevant or conflicting ones. Specifically, this mechanism computes pairwise task similarities between clients and assigns client-specific aggregation weights accordingly, so that each client can emphasize contributions from similar clients for group-wise knowledge sharing, while down-weighting the influence of unrelated or conflicting tasks. Additionally, to address the challenge of limited parameter variation in FedFM, we adopt task vector for aggregation inspired by prior work \cite{ilharco2022editing}. The task vector is computed as the difference between the local fine-tuned model and its initialization at each communication round, serving as a compact and informative representation of the client's task for effective similarity estimation even under constrained update spaces. In this way, FedBip effectively balances personalization and federation by adaptively aligning aggregation with task similarity, enabling both individual task adaptation and group-wise knowledge transfer. Extensive experiments on benchmarks demonstrate the effectiveness of our method. Our contribution can be summarized as:
\begin{itemize}
    \item A novel bi-level personalization framework tailored for federated foundation models, enabling personalized learning at both the client and server levels. 
    \item An innovative server-side personalization method based on task vector aggregation, enhancing global model adaptability across heterogeneous clients. 
    \item A new lightweight joint fine-tuning strategy for small-scale adaptation of pre-trained foundation models, improving efficiency and performance in federated settings. 
    \item A comprehensive empirical analysis demonstrating the effectiveness and robustness of the proposed framework across diverse benchmark datasets and scenarios.
\end{itemize}

\section{Related Work}

\subsection{Federated Foundation Model}
Federated foundation models \cite{zhuang2023foundation,yu2023federated,ren2024advances} have been proposed to adapt the capabilities of large-scale pre-trained models in FL settings while preserving data privacy and enabling collaborative learning across decentralized clients. FedIT \cite{zhang2023towards} serves as an early attempt to adapt FedAvg for large-scale foundation models. Based on this, a growing body of research has emerged to tackle various challenges in FedFM, including efficiency \cite{zhang2023fedpetuning,yang2025federated}, privacy \cite{han2024fedsecurity}, and heterogeneity \cite{cho2023heterogeneous,sun2024improving,wang2024flora}.
As the diversity of client data and tasks increases, personalization becomes a main concern in FedFM. Based on the underlying personalization techniques used, existing methods can be broadly categorized into two main types: (1) methods that introduce additional personalized modules \cite{chen2024feddat,yang2024dual}, enabling each client to adapt to its local data distribution, while still benefiting from shared global knowledge; (2) methods that explicitly decouple global and personalized components \cite{guo2024selective}, allowing for more flexible and resource-efficient control over global collaboration and local adaption. However, these approaches primarily focus on client-level personalization, often overlooking the potential degradation of personalization performance from uninformed server-side aggregation strategies. To fill this gap, our work explores bi-level personalization in FedFM by jointly considering both client-level and server-level personalization, extending the scope of research in this area.

\subsection{Aggregation-Weighting Methods in FL}
In conventional FL, there are also various studies exploring aggregation-weighting strategies to balance local adaptation and federation by adjusting the contribution of each client during aggregation. These approaches can be broadly categorized into two classes: global model aggregation-weighting methods and personalized aggregation-weighting methods. For global model aggregation-weighting methods, the goal is to mitigate inter-client divergence by assigning different aggregation weights across clients to reduce the impact of conflicting updates, thereby enhancing global model performance. For example, some approaches, such as FedDisco \cite{ye2023feddisco} and L-DAWA \cite{rehman2023dawa}, explicitly adjust the server’s aggregation weights by quantifying the divergence between the local and global model distributions or parameter spaces to mitigate inter-client divergence. Other methods like FedLAW \cite{li2023revisiting} and FedAWA \cite{shi2025fedawa} propose to learn the aggregation weights by optimizing the divergence between local client models and the global model, thereby enabling more adaptive aggregation. In contrast, personalized aggregation methods further enhance performance by employing client-specific aggregation-weighting strategies, enabling better alignment with local objectives. FedAMP \cite{huang2021personalized} introduces an attention-based mechanism that enables each client to perform personalized aggregation by weighing peer models based on parameter similarity. Similarly, pFedLA \cite{ma2022layer} employs a hypernetwork to learn a layer-wise aggregation policy, assigning distinct weights across layers to better personalize the aggregated model. Methods such as FedPHP \cite{li2021fedphp} and APPLE \cite{luo2022adapt} further aggregate client models locally with client-specific weights during local model updates. Additionally, approaches like PartialFed \cite{sun2021partialfed}, and FedALA \cite{zhang2023fedala} adopt adaptive strategies to blend the local and global models, thereby generating personalized models based on the extent of divergence or task relevance.
However, all these methods are based on conventional FL and do not consider the unique characteristics of FedFM, while our method is the first to leverage the capacity of pre-trained foundation models to realize the adaptive personalized aggregation-weighting for balanced personalization and federated learning in FedFM.

\begin{table*}
\begin{minipage}{\linewidth}
\begin{minipage}[h]{0.32\linewidth}
\makeatletter\def\@captype{figure}
\centering
\includegraphics[width=\linewidth]{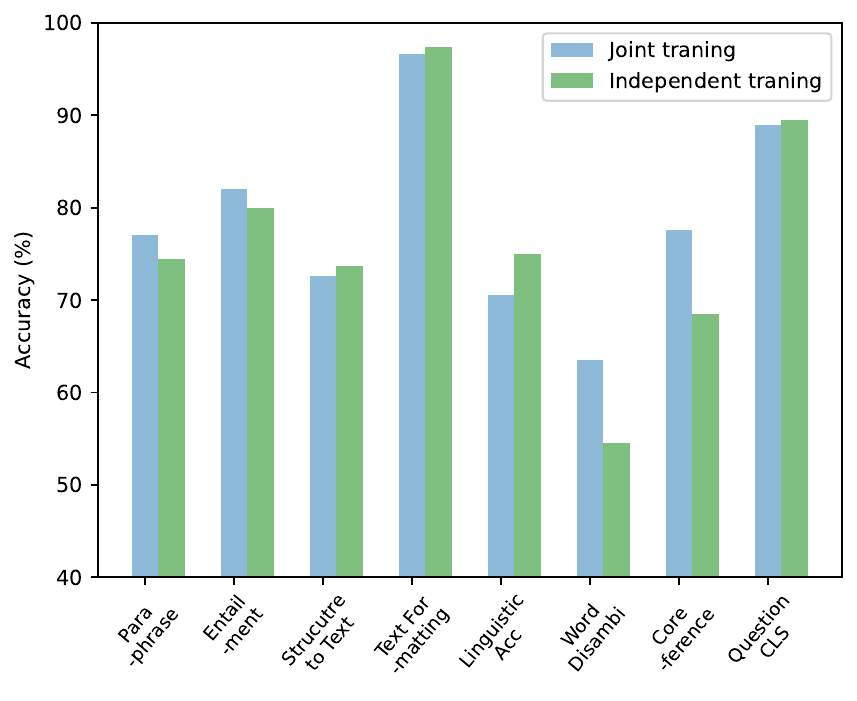} 
\vspace{-20pt}
\caption{\small Performance compared between jointly training and independently training.}
\label{fig-task-conflict}
\end{minipage}
\hfill
\begin{minipage}[h]{0.32\linewidth}
\makeatletter\def\@captype{figure}
\centering
\includegraphics[width=\linewidth]{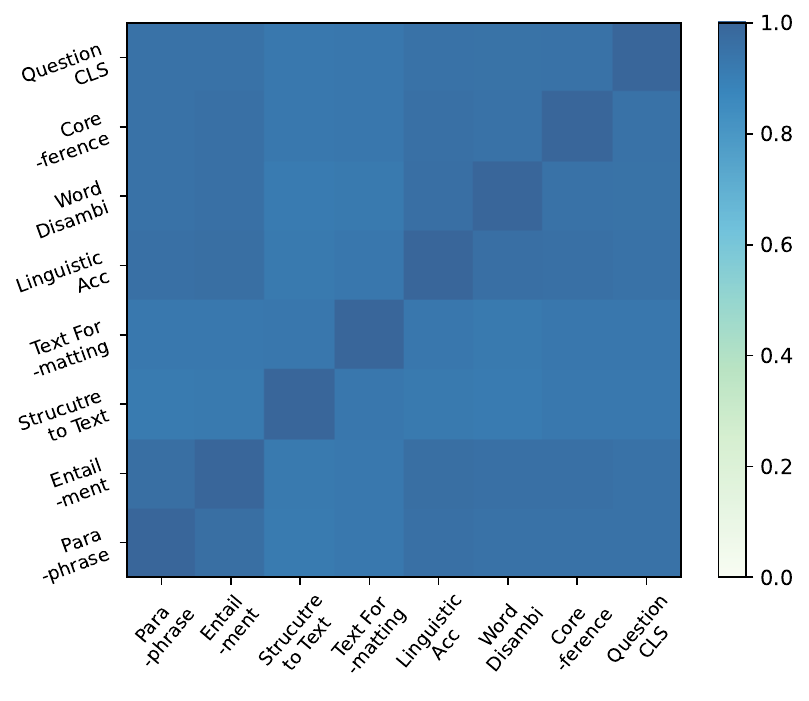} 
\vspace{-20pt}
\caption{\small Parameter similarity between models fine-tuned on different tasks.}
\label{fig-parameter-sim}
\end{minipage}
\hfill
\begin{minipage}[h]{0.32\linewidth}
\makeatletter\def\@captype{figure}
\centering
\includegraphics[width=\linewidth]{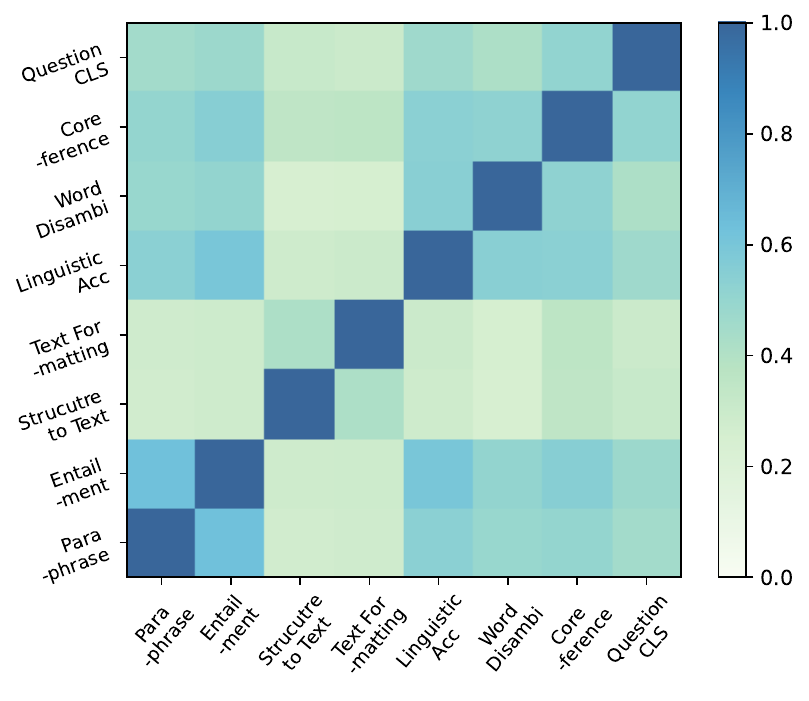} 
\vspace{-20pt}
\caption{\small Task vector similarity between models fine-tuned on different tasks.}
\label{fig-vector-sim}
\end{minipage}
\end{minipage}
\end{table*}

\subsection{Task Arithmetic}
Task Arithmetic \cite{ilharco2022editing} is proposed as an efficient model editing method for foundation models, enabling flexible adaptation and modification without full retraining. It introduces the concept of ``task vector'' as a compact representation of the transformation required to adapt a pre-trained model to a specific task through fine-tuning. By performing simple task vector arithmetic, it can create models with new capabilities for tasks. Several studies have explored the integration of task arithmetic into FL. For example, the study \cite{morafah2024towards} incorporates task arithmetic to merge knowledge distilled from heterogeneous models, while other work \cite{tao2024task} demonstrates the effectiveness of task arithmetic from a theoretical perspective of FL. However, none of the existing works have explored the use of task arithmetic in FedFM, and our work fills this gap by introducing task-vector aggregation tailored to the unique challenges of FedFM.

\section{Preliminary}
\subsection{Task Vector}
\label{sec-task-vector}
Task vector is introduced by \cite{ilharco2022editing} as an efficient method for model editing, which effectively encodes the necessary information required to perform a specific task and can serve as task-specific representations. Formally, given a pre-trained model $\theta_{pre}$ and its fine-tuned model $\theta^{i}_{ft}$ on task $i$, the task vector is obtained by the element-wise difference between the pre-trained and fine-tuned model, formulated as $\tau_{i}=\theta^{i}_{ft}-\theta_{pre}$. These task vectors can be used to transfer, combine, or remove task-specific knowledge through simple vector arithmetic. For example, by adding a task vector as $\theta_{pre}+\tau_i$, one can enable fast adaptation or transfer of the pre-trained model to task $i$.

\paragraph{Task Vector in PEFT.} To reduce computational overhead, parameter-efficient fine-tuning (PEFT) methods \cite{han2024parameter} have been proposed, which achieve efficient learning by updating only a small subset of model parameters while keeping the majority of parameters frozen. This can be formalized as $\theta=(\Delta\theta, \theta_{pre})$, where $\Delta\theta$ denotes the trainable subset and $\theta_{{pre}}$ remains fixed. Suppose the initial tunable parameters are $\Delta\theta^{0}$ and the fine-tuned parameters are $\Delta\theta^{t}$, then the task vector for task $i$ in the PEFT setting is $\tau_{i}=\Delta\theta^{i}-\Delta\theta^0$.

\subsection{Federated Foundation Models}
Federated foundation models \cite{zhuang2023foundation} are proposed to adapt large-scale foundation models within the FL framework by leveraging private data from distributed clients for model training. 
Given $K$ clients, each with its local dataset $D_k$, the overall process of FedFM is:
\begin{equation}
   \text{Client: } \theta_k = \min_{\theta} f_k(\theta;D_k), \quad \text{Server: } \theta = \sum_{k=1}^K p_k \theta_k, 
\end{equation}
where $f_k(\theta;D_k)$ is the local objective function computed on client $k$'s dataset $D_k$, and $p_k$ is the aggregated weight assigned to client $k$. As foundation models usually contains millions or even billions of parameters compared to traditional models, recent studies have incorporated PEFT method in FedFM for efficient learning, where the objective is to fine-tune only a small learnable subset of parameters, formulates as $\Delta\theta_k = \min_{\Delta\theta} f_k(\Delta\theta,\theta_{pre};D_k)$ on clients and $\Delta\theta = \sum_{k=1}^K p_k \Delta\theta_k$ on the server.



\section{Motivation}
As foundation models are pre-trained on large-scale and diverse data to acquire generalizable capabilities across various downstream tasks, the objective of FedFM fine-tuning is to adapt these generalized models to client-specific data in a privacy-preserving way, enabling better personalization for new users or scenarios. However, the unique characteristics of FedFM, such as high data heterogeneity and strong pre-trained initialization, introduce new challenges that go beyond those encountered in conventional FL. 

\paragraph{Personalization and Federation in FedFM.} 
Let $(\vx,\vy)$ denote a data pair sampled from the distribution $P$ and clients $i,j \in [K]$ and $i\neq j$.
In conventional FL, it primarily addresses heterogeneity with label shifts $P_i(\vy) \neq P_j(\vy)$ or feature shifts $P_i(\vx) \neq P_i(\vx)$. In contrast, FedFM often involves more complex heterogeneity, such as task shift or domain shift, denoted as $P_i(\vx,\vy) \neq P_j(\vx,\vy)$. This arises from the inherent nature of foundation models designed to support a wide range of tasks across diverse domains \cite{touvron2023llama}. 
Under such settings, conventional federation may lead to suboptimal performance, as it fails to account for task similarity and conflicts and treats all client contributions uniformly. This uniform aggregation can degrade personalized knowledge acquired from local fine-tuning, particularly when updates from irrelevant or conflicting tasks dominate, and hinder the exploitation of group-wise knowledge from similar tasks to enhance personalization.
To empirically examine this, we conduct an experimental analysis comparing two training setups: (1) jointly training all tasks, and (2) fine-tuning each task independently. As shown in Figure~\ref{fig-task-conflict}, we observe that for several tasks (e.g., Linguistic Acceptability), the independent performance is higher than that of joint training, indicating that inter-task conflicts exist under high heterogeneity. These findings suggest that naively aggregating all client models in conventional FL is suboptimal in such scenarios. Therefore, it is essential to develop new aggregation strategies that can account for task similarity and conflicts in FedFM.

\paragraph{Limited Parameter Variation in FedFM.} 
As demonstrated in prior work \cite{li2019convergence}, under standard assumptions of convexity and smoothness, the performance of a model trained for $T$ rounds can be bounded by $f(\theta_T)-f(\theta^*)\leq O(\frac{f(\theta_0)-f(\theta^*)}{T})$, where $\theta^*$ denotes the optimal model. In conventional FL, training typically begins from scratch, with random initialization $\theta_0=\gN(0,\sigma^2)$. In contrast, foundation models in FedFM are first pre-trained on large-scale datasets, then fine-tuning starts from the pre-trained weight $\theta_0 = \theta_{pre}$. Since pre-training imparts generalizable knowledge to provide a strong and generalized initialization for fine-tuning, the bound tends to be tighter for foundation models in FedFM compared to models trained from scratch in conventional FL, corresponding to relatively smaller model updates for optimization in FedFM. 
This limited variation between model parameters in FedFM poses significant challenges for adapting conventional FL methods, which typically rely on model parameter divergence to guide personalized aggregation and may fail to reflect meaningful task-level similarity or discrepancies, thereby undermining the effectiveness of these approaches. This issue becomes even more pronounced when using PEFT techniques, which tend to learn more similar parameters in a significantly lower-dimensional and restricted subspace. To empirically validate this,  we compute the cosine similarity between model parameters fine-tuned on different tasks. As shown in Figure~\ref{fig-parameter-sim}, the results reveal that parameter differences are insufficient to reflect the underlying task divergence, highlighting the need for novel methods in FedFM.

\section{Method}
\begin{figure}[t]
\centering
\includegraphics[width=0.5\textwidth]{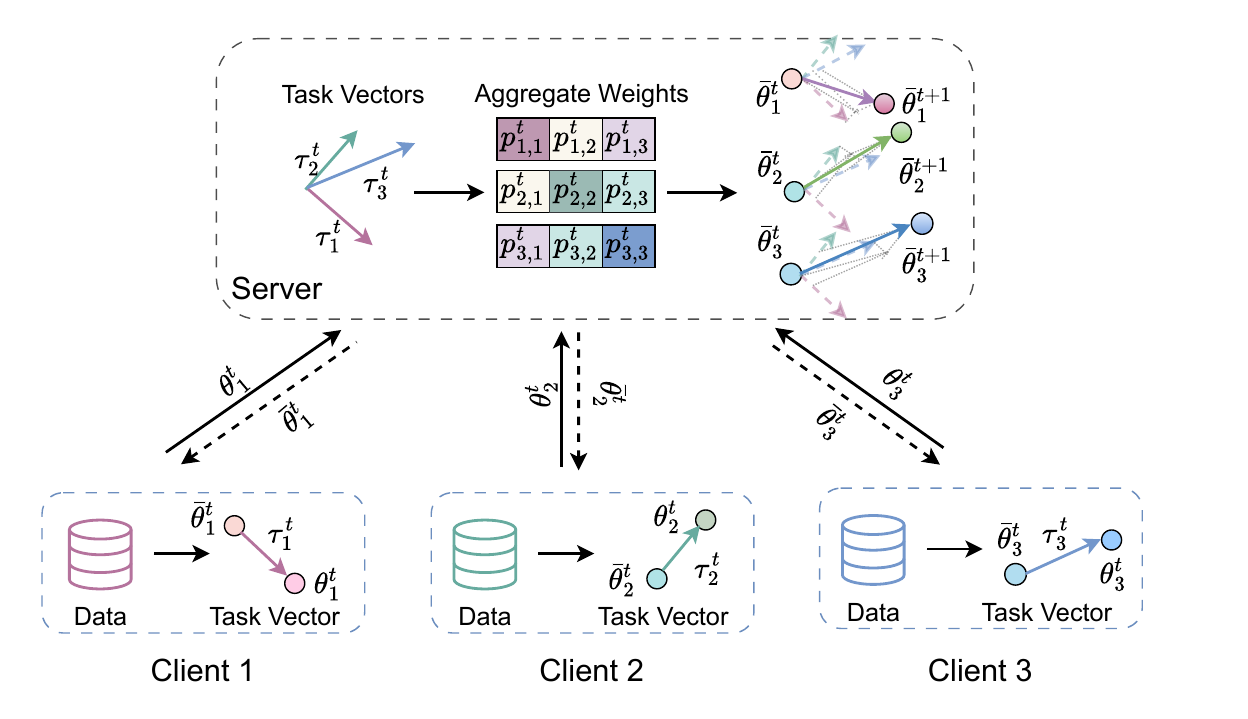} 
\caption{The overall framework of FedBip, consists of two components: client-level personalization, where each client fine-tunes the foundation model on its local data and computes a corresponding task vector; and server-level personalization, where the server performs personalized aggregation by computing task vector similarities and assigning client-specific aggregation weights to guide updates for each client.}
\label{fig-overall}
\end{figure}

As previously analyzed, conventional uniform federation may degrade the personalization performance in FedFM, due to the disturbance from irrelevant or conflicting tasks and the underutilization of group-wise knowledge from similar tasks. Additionally, the conventional FL methods for balancing personalization and federation often fail to generalize to FedFM, as they rely on parameter similarities across clients, which are often insufficient to capture meaningful task-level distinctions with limited parameter variation in FedFM. 
To address these challenges, we propose a bi-level personalization framework with task-vector aggregation, as shown in Figure~\ref{fig-overall}. During learning, each client performs local fine-tuning of the foundation model on its private dataset to achieve task-specific adaptation, representing the client-level personalization. Subsequently, the server conducts task-vector aggregation, computing client-specific aggregation weights based on inter-task relationships derived from task vectors for better personalization and federation balance, representing the server-level personalization. 

\subsection{Client-level Personalization}
For client-level personalization, the objective is to adapt the foundation model to the specific task of each client. To achieve this, we follow the standard federated learning paradigm, where each client fine-tunes the model on its local dataset to obtain a personalized model. More specifically, during each communication round $t$, each client $i$ receives its aggregated model $\bar{\theta}^t_i$ from the server as the initialization, and then fine-tunes it on its local dataset $D_i$ for $E$ local epochs, resulting in the personalized model $\theta_i^t$, which is subsequently sent back to the server.
\begin{equation}
    \theta_i^t = \min_{\theta} f_i(\theta;D_k), \quad \text{initialized with } \theta=\bar{\theta}^t_i
\end{equation}

\subsection{Server-level Personalization}
Since conventional federated aggregation may degrade the personalization performance achieved at the client level, we propose a task-vector aggregation strategy to better balance personalization and federation, while enabling group-wise knowledge sharing for server-level personalization. 

To balance personalization and federation, we seek to amplify the aggregation weights for clients with similar tasks, while down-weighting those from irrelevant or conflicting clients. Conventional FL methods often achieve this by using each client's model parameter as its representation to infer task similarity or divergence. However, in FedFM, due to the limited parameter variation, such approaches may fail as discussed in Motivation. To overcome this limitation, we propose using task vectors as more informative representations to guide aggregation weights, inspired by prior work \cite{ilharco2022editing}.
Since each client in FedFM typically holds a local dataset corresponding to a different task, the task vector between the locally fine-tuned model and the aggregated model can capture task-specific characteristics, even under the limited parameter variation imposed by FedFM. To empirically validate this, we compute the cosine similarity between task vectors obtained from models fine-tuned on different tasks. As illustrated in Figure~\ref{fig-vector-sim}, the results show that task vectors effectively reflect inter-task divergence, indicating their potential to serve as a signal for task-aware aggregation in FedFM.

Additionally, considering that the relationships between different tasks of clients are inherently non-uniform, we adopt a task-vector aggregation for each client instead of directly aggregating model parameters into a single global model. This approach offers two key advantages: (1) it enables more precise and adaptive updates by leveraging the most recent task-specific information for each client, thereby facilitating client-specific model updates; and (2) it mitigates the risk of error accumulation across communication rounds, as the aggregation focuses solely on the current round’s task vectors, rather than recursively averaging full model weights.
Moreover, since all clients fine-tune their models from a shared pre-trained initialization, the resulting task vectors reside in a common representation space, making it both meaningful and valid to perform vector arithmetic. As a result, the aggregated task vector can be reliably added to each client’s previously personalized aggregated model, ensuring consistent and effective model adaptation.

More speicialy, on the server side, task vectors are first computed for each client $i$ as:
\begin{equation}
\label{eq-vector}
\tau^t_i=\theta_i^t-\bar{\theta}_i^t,  \forall i \in [K],
\end{equation}
where $\bar{\theta}_i^t$ denotes the maintained aggregated model from the previous communication round, and the task vector $\tau^t_i$ captures the task-specific adaptation information of client $i$.
To perform task-aware aggregation, we compute a set of similarity-based weights for each client. For client $i$, the aggregation weight associated with each peer client $k\in\{1,...,K\}$ is computed as:
\begin{equation}
\label{eq-vector-agre}
p^t_{i,k} = \frac{g(\tau_i^t,\tau_k^t)}{\sum_{j=1}^K g(\tau_i^t,\tau_j^t)},
\end{equation}
where $g(\cdot,\cdot)$ is a similarity function, and the weights are normalized across all clients to ensure $\sum_{k=1}^Kp^t_{i,k}=1$. Here, we choose cosine similarity as it is invariant to vector magnitude.
Finally, the server computes a personalized aggregated model for client $i$ as $\bar{\theta}_i^{t+1} = \bar{\theta}_i^{t}+\sum_{k=1}^K p_{i,k}^t \tau_k^t$, which is sent back to client $i$ for next round. The overall process of FedBip is in Algorithm~\ref{alg:algorithm}

\begin{algorithm}[tb]
\caption{FedBip}
\label{alg:algorithm}
\textbf{Input}: Clients $K$, local datasets $\{D_1,...,D_K\}$, local epoch $E$, communication rounds $T$ \\
\textbf{Output}: Personalized models $\theta_1,...,\theta_K$
\begin{algorithmic}[1] 
\STATE Clients initialize local model
\FOR{$t=1,...,T$}
\STATE Server sends aggregated $\bar{\theta}^t_{0},...,\bar{\theta}^t_{K}$ to each client
\FOR{each client $i \in [K]$ in parallel}
\STATE $\theta_i^t \leftarrow $ClientUpdate($\bar{\theta}_i^t,D_k,E$)
\STATE Client $i$ sends $\theta_i^t$ to the server
\ENDFOR
\FOR{$i \in [K]$}
\IF{FedBip}
\STATE Server obtains $\tau^t_{1},...,\tau^t_{K}$ by Equation~\ref{eq-vector}
\STATE Server obtains $p_{i,1}^t,...,p_{i,K}^t$ by Equation~\ref{eq-vector-agre}
\STATE Server aggregates $\bar{\theta}_i^{t+1} = \bar{\theta}_i^{t}+\sum_{k=1}^K p_{i,k}^t \tau_k^t$
\ENDIF
\IF{FedBip-L}
\STATE Server obtains $\tau^t_{1,l},...,\tau^t_{K,l}$ by Equation~\ref{eq-vectorl}
\STATE Server obtains $p_{i,1,l}^t,...,p_{i,K,l}^t$ by Equation~\ref{eq-vector-agrel}
\STATE Server aggregates $\bar{\theta}_{i,l}^{t+1} = \bar{\theta}_{i,l}^{t}+\sum_{k=1}^K p_{i,k,l}^t \tau_{k,l}^t$ 
\ENDIF
\ENDFOR
\ENDFOR
\STATE \textbf{return} $\theta_1,...,\theta_K$
\end{algorithmic}
\end{algorithm}

\paragraph{Layer-Wise Extension of FedBip}
Previous studies \cite{ma2022layer,rehman2023dawa,lee2023layer} have demonstrated that model divergence often varies across different layers, highlighting layer-wise heterogeneity as a vital factor in FL. This phenomenon is also pronounced in foundation models, where different layers serve distinct roles—for instance, lower layers in transformer-based models tend to encode local features, while higher layers capture more abstract information \cite{chefer2021transformer}. Motivated by this, we extend FedBip to support layer-wise task-vector aggregation, enabling finer-grained adjustment over the aggregation process. Specifically, for each communication round $t$, and each layer $l$, we compute the layer-wise task vector and corresponding aggregation weights as follows:   
\begin{align}
\label{eq-vectorl}
\tau^t_{i,l} & =\theta_{i,l}^t-\bar{\theta}_{i,l}^t, \\
\label{eq-vector-agrel}
p_{i,k,l}^t & = \frac{g(\tau_{i,l}^t,\tau_{k,l}^t)}{\sum_{j=1}^K g(\tau_{i,l}^t,\tau_{j,l}^t)},
\end{align}
where $\theta_{i,l}^t$ denotes the parameters of the $l$-th layer of the fine-tuned model from client $i$ and $\bar{\theta}_{i,l}^t$ denotes  the parameters of the $l$-th layer of the previous aggregated model for client $i$. The server then performs layer-wise task-vector aggregation for client $i$ as $\bar{\theta}_{i,l}^{t+1} = \bar{\theta}_{i,l}^{t}+\sum_{k=1}^K p_{i,k,l}^t \tau_{k,l}^t$.

\paragraph{Remark.}
The proposed aggregation algorithm is modular and flexible, and can be seamlessly integrated with other federated client-side optimization techniques such as FedProx \cite{li2020federated} to further enhance performance. Moreover, it is also applicable to FedFM settings with PEFT technologies. In such cases, the task vectors can be computed using only the trainable parameters, as described in Section Preliminary, without modification to the overall framework.

\definecolor{mygray}{gray}{0.95}
\begin{table*}[h]
    \small
    \setlength{\tabcolsep}{4pt}
    \centering
    \begin{tabular}{l|cccccccc|c|cc}
        \toprule[1.25pt]
        \multirow{1}{*}{Methods} 
        &\multicolumn{1}{l}{\makecell[c]{Para \\ -phrase}}  &\multicolumn{1}{l}{\makecell[c]{Entail \\ -ment}}  &\multicolumn{1}{l}{\makecell[c]{Structure \\ to Text}} &\multicolumn{1}{l}{\makecell[c]{Text For \\ -matting}} &\multicolumn{1}{l}{\makecell[c]{Linguistic \\ Acc}}  &\multicolumn{1}{l}{\makecell[c]{Word \\ Dis}} &\multicolumn{1}{l}{\makecell[c]{Core \\ -ference}} &\multicolumn{1}{l}{\makecell[c]{Question \\ CLS}} &\multicolumn{1}{l}{\makecell[c]{Average}} & \multicolumn{1}{|l}{ \makecell[c]{Client \\ Comp.over}} & \makecell[c]{Server \\ Exe.time}
\\ \hline 
FedIT   &77.00  &82.50  &71.99  &96.61  &76.00 &62.00 &73.26 &83.00 &79.05 & 0.281 TFLOPS & 0.1s\\
FedAWA         &70.00  &85.50  &71.55  &96.82  &75.50 &63.00 &73.86 &93.50 &78.72 & 0.281 TFLOPS& 1.3s\\
L-DAWA       &74.00  &83.00  &70.57  &\textbf{96.97}  &75.50 &64.50 &76.47 &92.50 &79.19 &0.281 TFLOPS & 0.2s\\
\hline
FedAMP &\textbf{83.50}  &82.50  &72.30  &96.96  &78.50 &52.50 &76.60 &93.50 &79.55 &0.281 TFLOPS & 2.2s \\
FedALA &75.00 &84.50 &71.13 &96.51 &76.00 &\textbf{67.40} &76.6 &90.50 &79.71 &0.562 TFLOPS & 0.1s\\
\hline
FedBip &80.00 &85.00  &72.20  &96.74  &79.50 &{60.50} &\textbf{82.20} &93.50 &81.21 &0.281 TFLOPS & 2.4s \\
 FedBip-L     &{81.00}  &\textbf{87.00}  &\textbf{73.23}  &96.61  &\textbf{80.50} &62.00 &{79.89} &\textbf{95.00} &\textbf{81.90} &0.281 TFLOPS & 4.8s\\

\hline 
 \rowcolor{mygray}\multicolumn{12}{l}{ \emph{Modularity}} \\
FFA-LoRA &62.50 &60.00 &63.10 &96.30 &72.00 &52.00 &\textbf{61.83} &74.50 &67.78 &0.281 TFLOPS & 0.1s \\
+FedBip &\textbf{72.50} &\textbf{74.50} &\textbf{69.42} &\textbf{96.40} &\textbf{79.00} &\textbf{53.75} &58.56 &\textbf{86.00} &\textbf{73.77} &0.281 TFLOPS & 2.4s \\
\hline
FedDPA &80.00 &\textbf{87.50} &73.38 &96.74 &78.50 &\textbf{64.00} &\textbf{79.84} &94.00 &81.74 &0.562 TFLOPS & 0.1s\\
+FedBip &\textbf{84.50} &86.50 &\textbf{73.76} &\textbf{96.81} &\textbf{78.50} &62.00 &79.52 &\textbf{94.50} &\textbf{82.01} &0.562 TFLOPS & 2.4s\\

        \bottomrule[1.25pt]
    \end{tabular}
    \caption{Results of different models on NLP tasks with modularity and efficiency results.}
    \label{table-nlp}
\end{table*}

\begin{table*}
\begin{minipage}{\linewidth}
\begin{minipage}[h]{0.45\linewidth}
\makeatletter\def\@captype{table}
 \small
    \setlength{\tabcolsep}{4pt}
    \centering
    \begin{tabular}{l|cccc|c}
        \toprule[1.25pt]
        {Methods} 
        &\multicolumn{1}{l}{\makecell[c]{Art }}  &\multicolumn{1}{l}{\makecell[c]{CliPart  }}  &\multicolumn{1}{l}{\makecell[c]{Product}} &\multicolumn{1}{l}{\makecell[c]{Real World}}  &\multicolumn{1}{l}{\makecell[c]{Average}}
\\ \hline 
FedAVG   & 84.71& 86.36& 94.44& 94.26& 90.02 \\
FedAWA   & 86.36 & 86.09& 94.31& 94.46 & 90.31 \\
L-DAWA   & 84.43 & 85.98 & 94.37& 94.81 & 89.90 \\
\hline
FedAMP & 87.99 & 90.11 & 96.18& 95.84 & 92.53 \\
FedALA & \textbf{89.64} & 90.89 & 96.18& 96.39 & 93.33  \\
\hline
FedBip &88.56 & 91.10 & \textbf{97.97} & 96.01 & 93.41 \\
 FedBip-L & 88.02 & \textbf{91.62} & 97.94 & \textbf{97.39} & \textbf{93.74}\\

\hline 
 \rowcolor{mygray}\multicolumn{6}{l}{ \emph{Modularity}} \\
FedProx & 84.15 & 86.17 & 93.95 & 94.43 & 89.68 \\
 + FedBip & \textbf{90.19} & \textbf{92.30} & \textbf{98.13} & \textbf{97.15} & \textbf{94.44} \\
 \hline
 Ditto & 87.19 & 90.36 & 96.84 & 95.19 & 92.39 \\
+FedBip & \textbf{88.02} & \textbf{91.96} & \textbf{97.72} & \textbf{96.84} & \textbf{93.63} \\

        \bottomrule[1.25pt]
    \end{tabular}
    \caption{\small Results of different models on CV tasks.}
    \label{table-cv}
\end{minipage}
\hfill
\begin{minipage}[h]{0.26\linewidth}
\makeatletter\def\@captype{figure}
\centering
\includegraphics[width=\linewidth]{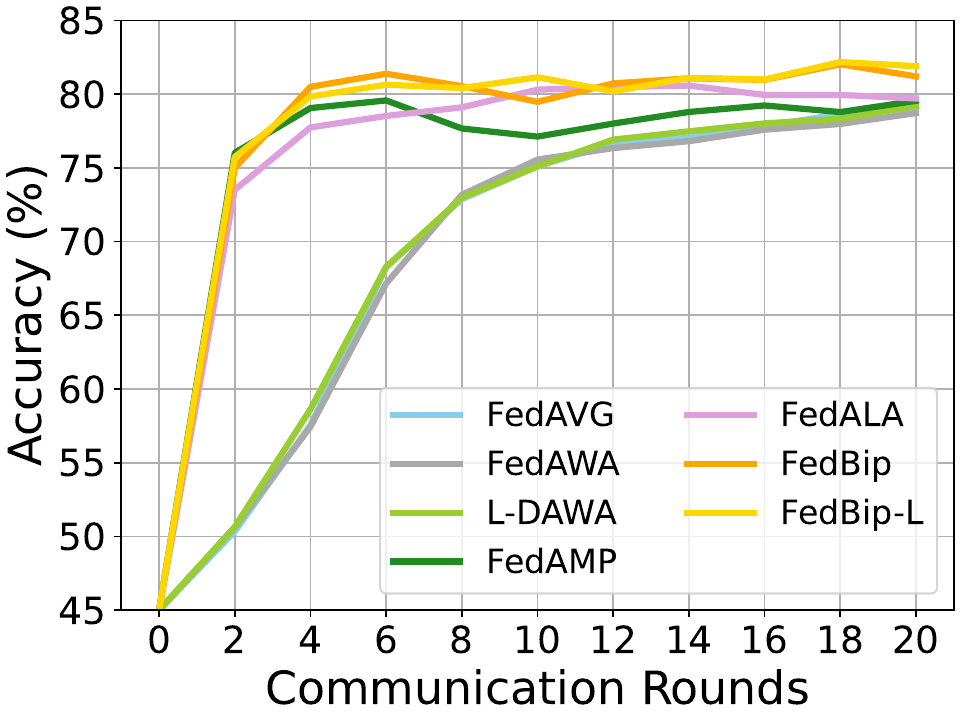} 
\caption{\small Accuracy via communication rounds in NLP domain.}
\label{fig-conv-nlp}
\end{minipage}
\hfill
\begin{minipage}[h]{0.26\linewidth}
\makeatletter\def\@captype{figure}
\centering
\includegraphics[width=\linewidth]{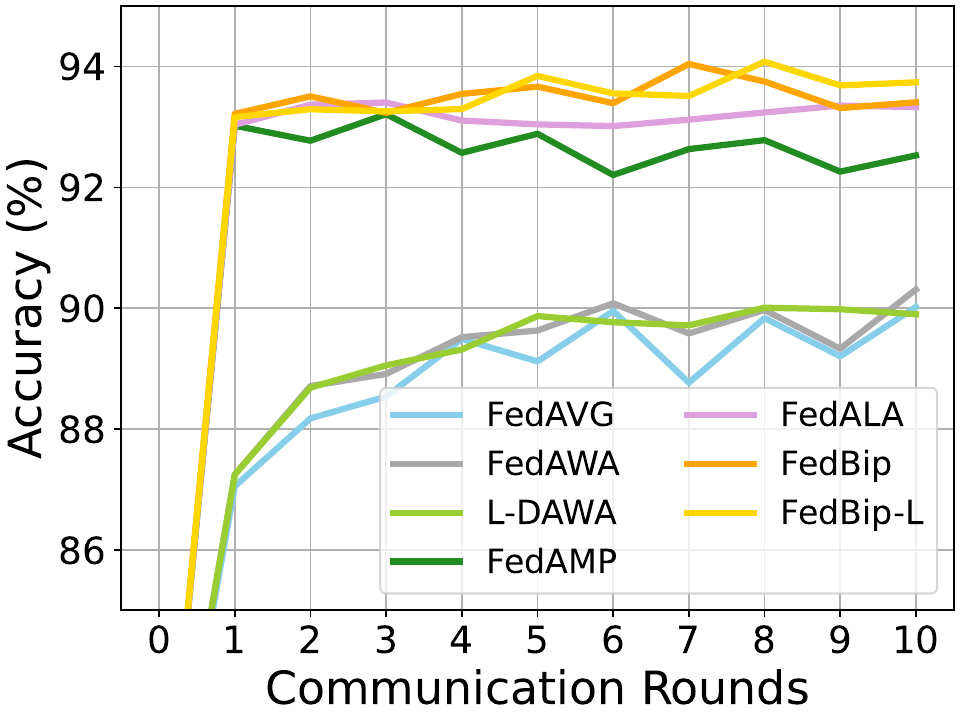} 
\caption{\small Accuracy via communication rounds in CV domain.}
\label{fig-conv-cv}
\end{minipage}
\end{minipage}
\end{table*}

\section{Experiment}
\subsection{Experiment Setting}
\paragraph{Datasets.} 
To comprehensively evaluate the effectiveness of FedBip, we conduct experiments in both the computer vision (CV) and natural language processing (NLP) domains.
For CV, we use the OfficeHome dataset \cite{venkateswara2017deep}, which comprises images across four distinct domains with 65 categories to simulate cross-domain heterogeneity. For NLP, we utilize the Flan \cite{wei2021finetuned}, which contains a diverse collection of instruction-following datasets, and we select eight distinct tasks as the federated dataset to simulate cross-task heterogeneity.

\paragraph{Baselines and Implementation.} 
We compare our methods with below baselines based on the same model architecture: 1) conventional global aggregation methods: FedAVG \cite{mcmahan2017communication} for CV and FedIT \cite{zhang2023towards} for NLP; 2) global model aggregation adjustment methods: FedAWA \cite{shi2025fedawa} and L-DAWA \cite{rehman2023dawa}; 3) personalized aggregation methods: FedAMP \cite{huang2021personalized} and FedALA \cite{zhang2023fedala}.
To simulate data heterogeneity, we distribute clients based on domain (in CV) or task (in NLP). For both settings, to better evaluate the effectiveness of methods, we assume that all clients are activated for every communication round and set round $T=10$ for CV and $T=20$ for NLP. For CV, we use the ViT backbone from the CLIP model \cite{radford2021learning} and apply full fine-tuning. For NLP, we employ LoRA \cite{hu2022lora} as the PEFT method for LLaMA-7B\footnote{https://huggingface.co/huggyllama/llama-7b}. 

\begin{table*}
\begin{minipage}{\linewidth}
\begin{minipage}[h]{0.28\linewidth}
\makeatletter\def\@captype{table}
 \small
    \setlength{\tabcolsep}{4pt}
    \centering
    \begin{tabular}{l|ccc}
        \toprule
        {Client Number} 
        &8  & 24 & 40
\\ \hline 
FedIT   & 79.05 & 65.07 & 66.67 \\
FedAWA &78.72 & 64.69 &66.49 \\
L-DAWA & 79.19&64.64 &65.78 \\
\hline
FedAMP &79.55 &66.37 &68.09 \\
FedALA &79.71 &67.24 &71.53 \\
\hline
FedBip &\textbf{81.21} &\textbf{68.91} &\textbf{73.00} \\
        \bottomrule
    \end{tabular}
\caption{\small Ablation study of different client numbers.}
\label{table-scale}
\end{minipage}
\hfill
\begin{minipage}[h]{0.71\linewidth}
\makeatletter\def\@captype{figure}
\centering
\includegraphics[width=\linewidth]{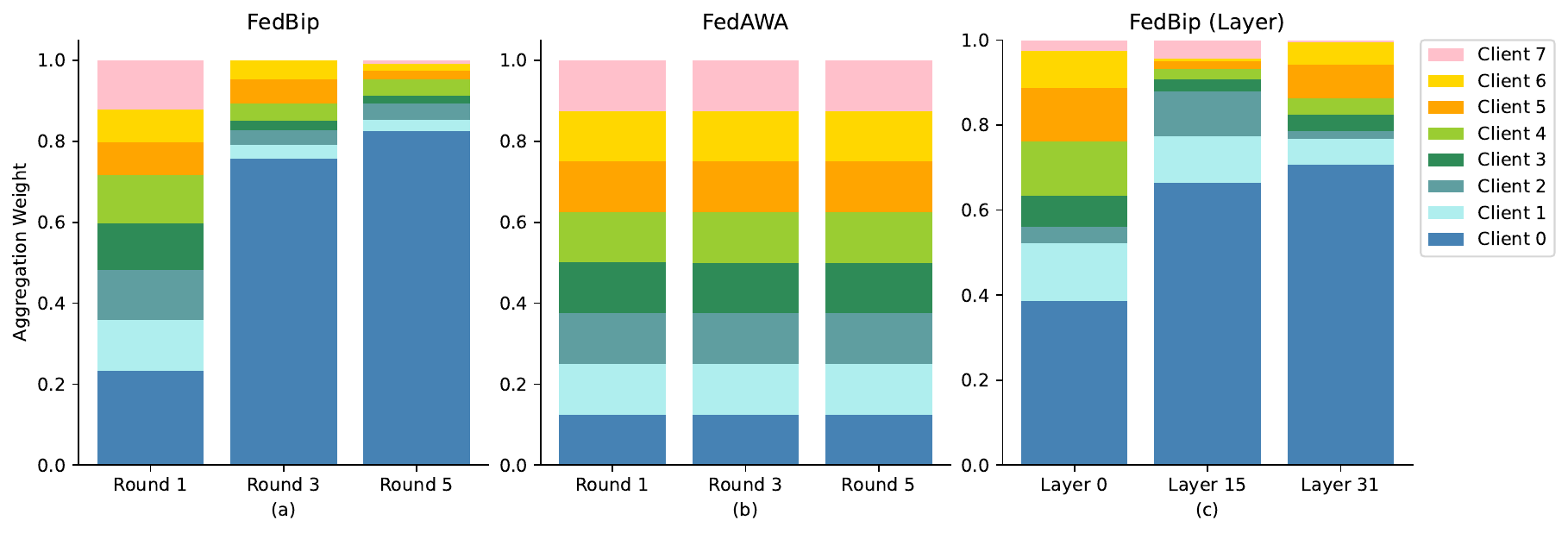} 
\vspace{-20pt}
\caption{\small Client 0's aggregation weights calculated by FedBip, FedAWA and FedBip-L.}
\label{fig-weight}
\end{minipage}
\end{minipage}
\end{table*}

\subsection{Main Results}
We evaluate FedBip against other baselines under two settings: full fine-tuning on cross-domain heterogeneity in CV and parameter-efficient fine-tuning on cross-task heterogeneity in NLP. As shown in Table~\ref{table-nlp} and Table~\ref{table-cv}, FedBip consistently achieves the best performance on average, demonstrating the effectiveness of our method for considering both client-level and server-level personalization. Moreover, the layer-wise extension (FedBip-L) further improves performance over FedBip, highlighting the importance of capturing layer-specific differences across clients. In addition, personalized aggregation methods generally outperform global aggregation baselines, emphasizing the necessity of aligning the model with client-specific distributions to handle heterogeneous tasks or domains. In particular, for tasks such as Linguistic Acceptability, where independent fine-tuning outperforms joint training (as shown in Figure~\ref{fig-task-conflict}), our method significantly improves performance. This provides further empirical evidence that FedBip effectively balances the federation and personalization in FedFM. 

\paragraph{Modularity.} 
FedBip is designed as a modular framework that can be seamlessly integrated with various existing client-side optimization FL algorithms to enhance their performance. As shown in Table~\ref{table-nlp} and Table~\ref{table-cv}, we integrate FedBip with several representative FL baselines. In the NLP setting, we combine FedBip with LoRA-based methods such as FFA-LoRA \cite{sun2024improving} and FedDPA \cite{yang2024dual}; in the CV setting, we combine it with both global method FedProx \cite{li2020federated} and personalized method Ditto \cite{li2021ditto}. The results show consistent average performance gains across all combinations, demonstrating the strong modularity and generalizability of our approach.

\paragraph{Flexibility.} 
FedBip is highly flexible and can be readily applied to different transformer-based foundation models as well as PEFT techniques. As illustrated in Table~\ref{table-nlp} and Table~\ref{table-cv}, FedBip is successfully deployed across models of varying scales and domains, including ViT (for CV) and LLaMA (for NLP). It also performs well under both full fine-tuning and LoRA-based PEFT, further highlighting its adaptability. The consistent performance improvements over baseline methods confirm that FedBip is a flexible and effective framework for addressing personalization in FedFM.

\subsection{Analysis}
\paragraph{Convergence Analysis.}
To analyze convergence, we compare average test accuracy versus communication rounds in Figure~\ref{fig-conv-nlp} and Figure~\ref{fig-conv-cv}. The results show that personalized aggregation methods, including FedBip, converge faster than global aggregation methods and reach stable performance after only a few rounds. Moreover, our method consistently outperforms all baselines throughout the training process.

\paragraph{Efficiency Analysis.} We evaluate the computation cost on the client side and execution time on the server side. As shown in Table~\ref{table-nlp}, FedBip introduces no additional computation overhead for clients and only a slight increase in server-side execution time, demonstrating that FedBip is a lightweight and efficient framework for FedFM fine-tuning.

\paragraph{Ablation Analysis.} To evaluate scalability, we vary client number $K \in \{8, 24, 40\}$ in Table~\ref{table-scale}, where FedBip consistently outperforms other baselines, demonstrating its effectiveness and robustness under increasing client heterogeneity and numbers. To evaluate impact of $g(\cdot,\cdot)$ in task-vector aggregation, we compare different similarity metrics. 
\begin{table}[h]
    \setlength{\tabcolsep}{4pt}
    \centering
    \begin{tabular}{l|ccc}
        \toprule
        {Similarity Metric} 
        &L2  & Pearson & Cosine
\\ \hline 
FedBip   & 79.55 & 80.77 & 81.21 \\
        \bottomrule
    \end{tabular}
    \caption{\small Ablation of similarity metric.}
    \label{table-abalation-similarity}
\end{table}

\begin{table}[h]
    \setlength{\tabcolsep}{4pt}
    \centering
    \begin{tabular}{l|ccc}
        \toprule
        Weighting 
        & Parameter & Parameter & Vector \\
        Aggregation 
        & Parameter & Vector & Vector
\\ \hline 
   & 79.63& 80.15& 81.21\\
        \bottomrule
    \end{tabular}
    \caption{\small Ablation of weighting and aggregation strategies.}
    \label{table-abalation-vector}
\end{table}
As shown in Table~\ref{table-abalation-similarity}, cosine similarity performs best, likely due to its magnitude invariance and better robustness for high-dimensional vectors. Additionally, we evaluate different weighting and aggregation strategies (Table~\ref{table-abalation-vector}). Results show that computing aggregation weights and performing aggregation directly on task vectors yields the best performance. This aligns with our earlier analysis that parameter-based similarity fails to capture task divergence, and parameter aggregation may lead to error accumulation.

\paragraph{Aggregation Weight Analysis.} We analyze the aggregation weights computed by FedBip, FedAWA, and FedBip-L in Figure~\ref{fig-weight}. (a) shows that weights of FedBip increasingly favor personalization over rounds, and FedBip effectively assigns higher weights to similar tasks while reducing influence of irrelevant ones compared with (b). (c) illustrates the variation in layer-wise aggregation weights of FedBip-L, indicating that lower layers share more common knowledge, whereas higher layers retain more personalized information.

\section{Conclusion}
FedFM fine-tuning aims to adapt pre-trained foundation models for a small group of new users across distributed clients, where balancing federation and personalization presents a critical challenge. To address this, we propose FedBip, a bi-level personalization framework that incorporates client-level personalization through local fine-tuning and server-level personalization via task-vector aggregation. Experimental results on both CV and NLP benchmarks demonstrate the effectiveness of FedBip, paving the way for future exploration of more advanced methods across diverse modalities and large-scale federated learning scenarios.




\bibliography{aaai2026}

\begin{thebibliography}{38}
\providecommand{\natexlab}[1]{#1}

\bibitem[{Chefer, Gur, and Wolf(2021)}]{chefer2021transformer}
Chefer, H.; Gur, S.; and Wolf, L. 2021.
\newblock Transformer interpretability beyond attention visualization.
\newblock In \emph{Proceedings of the IEEE/CVF conference on computer vision and pattern recognition}, 782--791.

\bibitem[{Chen et~al.(2024)Chen, Zhang, Krompass, Gu, and Tresp}]{chen2024feddat}
Chen, H.; Zhang, Y.; Krompass, D.; Gu, J.; and Tresp, V. 2024.
\newblock Feddat: An approach for foundation model finetuning in multi-modal heterogeneous federated learning.
\newblock In \emph{Proceedings of the AAAI Conference on Artificial Intelligence}, volume~38, 11285--11293.

\bibitem[{Cho et~al.(2023)Cho, Liu, Xu, Fahrezi, Barnes, and Joshi}]{cho2023heterogeneous}
Cho, Y.~J.; Liu, L.; Xu, Z.; Fahrezi, A.; Barnes, M.; and Joshi, G. 2023.
\newblock Heterogeneous lora for federated fine-tuning of on-device foundation models.
\newblock In \emph{International Workshop on Federated Learning in the Age of Foundation Models in Conjunction with NeurIPS 2023}.

\bibitem[{Guo et~al.(2024)Guo, Zeng, Wang, Fan, Wang, and Qu}]{guo2024selective}
Guo, P.; Zeng, S.; Wang, Y.; Fan, H.; Wang, F.; and Qu, L. 2024.
\newblock Selective Aggregation for Low-Rank Adaptation in Federated Learning.
\newblock \emph{arXiv preprint arXiv:2410.01463}.

\bibitem[{Han et~al.(2024{\natexlab{a}})Han, Buyukates, Hu, Jin, Jin, Sun, Wang, Wu, Xie, Yao et~al.}]{han2024fedsecurity}
Han, S.; Buyukates, B.; Hu, Z.; Jin, H.; Jin, W.; Sun, L.; Wang, X.; Wu, W.; Xie, C.; Yao, Y.; et~al. 2024{\natexlab{a}}.
\newblock Fedsecurity: A benchmark for attacks and defenses in federated learning and federated llms.
\newblock In \emph{Proceedings of the 30th ACM SIGKDD Conference on Knowledge Discovery and Data Mining}, 5070--5081.

\bibitem[{Han et~al.(2024{\natexlab{b}})Han, Gao, Liu, Zhang, and Zhang}]{han2024parameter}
Han, Z.; Gao, C.; Liu, J.; Zhang, J.; and Zhang, S.~Q. 2024{\natexlab{b}}.
\newblock Parameter-efficient fine-tuning for large models: A comprehensive survey.
\newblock \emph{arXiv preprint arXiv:2403.14608}.

\bibitem[{Hu et~al.(2022)Hu, Shen, Wallis, Allen-Zhu, Li, Wang, Wang, Chen et~al.}]{hu2022lora}
Hu, E.~J.; Shen, Y.; Wallis, P.; Allen-Zhu, Z.; Li, Y.; Wang, S.; Wang, L.; Chen, W.; et~al. 2022.
\newblock Lora: Low-rank adaptation of large language models.
\newblock \emph{ICLR}, 1(2): 3.

\bibitem[{Huang et~al.(2021)Huang, Chu, Zhou, Wang, Liu, Pei, and Zhang}]{huang2021personalized}
Huang, Y.; Chu, L.; Zhou, Z.; Wang, L.; Liu, J.; Pei, J.; and Zhang, Y. 2021.
\newblock Personalized cross-silo federated learning on non-iid data.
\newblock In \emph{Proceedings of the AAAI conference on artificial intelligence}, volume~35, 7865--7873.

\bibitem[{Ilharco et~al.(2022)Ilharco, Ribeiro, Wortsman, Gururangan, Schmidt, Hajishirzi, and Farhadi}]{ilharco2022editing}
Ilharco, G.; Ribeiro, M.~T.; Wortsman, M.; Gururangan, S.; Schmidt, L.; Hajishirzi, H.; and Farhadi, A. 2022.
\newblock Editing models with task arithmetic.
\newblock \emph{arXiv preprint arXiv:2212.04089}.

\bibitem[{Lee, Zhang, and Avestimehr(2023)}]{lee2023layer}
Lee, S.; Zhang, T.; and Avestimehr, A.~S. 2023.
\newblock Layer-wise adaptive model aggregation for scalable federated learning.
\newblock In \emph{Proceedings of the AAAI Conference on Artificial Intelligence}, volume~37, 8491--8499.

\bibitem[{Li et~al.(2021{\natexlab{a}})Li, Hu, Beirami, and Smith}]{li2021ditto}
Li, T.; Hu, S.; Beirami, A.; and Smith, V. 2021{\natexlab{a}}.
\newblock Ditto: Fair and robust federated learning through personalization.
\newblock In \emph{International conference on machine learning}, 6357--6368. PMLR.

\bibitem[{Li et~al.(2020)Li, Sahu, Zaheer, Sanjabi, Talwalkar, and Smith}]{li2020federated}
Li, T.; Sahu, A.~K.; Zaheer, M.; Sanjabi, M.; Talwalkar, A.; and Smith, V. 2020.
\newblock Federated optimization in heterogeneous networks.
\newblock \emph{Proceedings of Machine learning and systems}, 2: 429--450.

\bibitem[{Li et~al.(2019)Li, Huang, Yang, Wang, and Zhang}]{li2019convergence}
Li, X.; Huang, K.; Yang, W.; Wang, S.; and Zhang, Z. 2019.
\newblock On the convergence of fedavg on non-iid data.
\newblock \emph{arXiv preprint arXiv:1907.02189}.

\bibitem[{Li et~al.(2021{\natexlab{b}})Li, Zhan, Shao, Li, and Song}]{li2021fedphp}
Li, X.-C.; Zhan, D.-C.; Shao, Y.; Li, B.; and Song, S. 2021{\natexlab{b}}.
\newblock Fedphp: Federated personalization with inherited private models.
\newblock In \emph{Joint European Conference on Machine Learning and Knowledge Discovery in Databases}, 587--602. Springer.

\bibitem[{Li et~al.(2023)Li, Lin, Shang, and Wu}]{li2023revisiting}
Li, Z.; Lin, T.; Shang, X.; and Wu, C. 2023.
\newblock Revisiting weighted aggregation in federated learning with neural networks.
\newblock In \emph{International Conference on Machine Learning}, 19767--19788. PMLR.

\bibitem[{Luo and Wu(2022)}]{luo2022adapt}
Luo, J.; and Wu, S. 2022.
\newblock Adapt to adaptation: Learning personalization for cross-silo federated learning.
\newblock In \emph{IJCAI: proceedings of the conference}, volume 2022, 2166.

\bibitem[{Ma et~al.(2022)Ma, Zhang, Guo, and Xu}]{ma2022layer}
Ma, X.; Zhang, J.; Guo, S.; and Xu, W. 2022.
\newblock Layer-wised model aggregation for personalized federated learning.
\newblock In \emph{Proceedings of the IEEE/CVF conference on computer vision and pattern recognition}, 10092--10101.

\bibitem[{McMahan et~al.(2017)McMahan, Moore, Ramage, Hampson, and y~Arcas}]{mcmahan2017communication}
McMahan, B.; Moore, E.; Ramage, D.; Hampson, S.; and y~Arcas, B.~A. 2017.
\newblock Communication-efficient learning of deep networks from decentralized data.
\newblock In \emph{Artificial intelligence and statistics}, 1273--1282. PMLR.

\bibitem[{Morafah et~al.(2024)Morafah, Kungurtsev, Chang, Chen, and Lin}]{morafah2024towards}
Morafah, M.; Kungurtsev, V.; Chang, H.; Chen, C.; and Lin, B. 2024.
\newblock Towards diverse device heterogeneous federated learning via task arithmetic knowledge integration.
\newblock \emph{Advances in Neural Information Processing Systems}, 37: 127834--127877.

\bibitem[{Radford et~al.(2021)Radford, Kim, Hallacy, Ramesh, Goh, Agarwal, Sastry, Askell, Mishkin, Clark et~al.}]{radford2021learning}
Radford, A.; Kim, J.~W.; Hallacy, C.; Ramesh, A.; Goh, G.; Agarwal, S.; Sastry, G.; Askell, A.; Mishkin, P.; Clark, J.; et~al. 2021.
\newblock Learning transferable visual models from natural language supervision.
\newblock In \emph{International conference on machine learning}, 8748--8763. PmLR.

\bibitem[{Rehman et~al.(2023)Rehman, Gao, De~Gusm{\~a}o, Alibeigi, Shen, and Lane}]{rehman2023dawa}
Rehman, Y. A.~U.; Gao, Y.; De~Gusm{\~a}o, P. P.~B.; Alibeigi, M.; Shen, J.; and Lane, N.~D. 2023.
\newblock L-dawa: Layer-wise divergence aware weight aggregation in federated self-supervised visual representation learning.
\newblock In \emph{Proceedings of the IEEE/CVF international conference on computer vision}, 16464--16473.

\bibitem[{Ren et~al.(2024)Ren, Yu, Peng, Tang, Li, Gao, Tan, Zhao, Li, Li et~al.}]{ren2024advances}
Ren, C.; Yu, H.; Peng, H.; Tang, X.; Li, A.; Gao, Y.; Tan, A.~Z.; Zhao, B.; Li, X.; Li, Z.; et~al. 2024.
\newblock Advances and Open Challenges in Federated Learning with Foundation Models.
\newblock \emph{arXiv preprint arXiv:2404.15381}.

\bibitem[{Shi et~al.(2025)Shi, Zhao, Zhang, Zhou, Guo, and Chang}]{shi2025fedawa}
Shi, C.; Zhao, H.; Zhang, B.; Zhou, M.; Guo, D.; and Chang, Y. 2025.
\newblock FedAWA: Adaptive Optimization of Aggregation Weights in Federated Learning Using Client Vectors.
\newblock In \emph{Proceedings of the Computer Vision and Pattern Recognition Conference}, 30651--30660.

\bibitem[{Sun et~al.(2021)Sun, Huo, Yang, and Bai}]{sun2021partialfed}
Sun, B.; Huo, H.; Yang, Y.; and Bai, B. 2021.
\newblock Partialfed: Cross-domain personalized federated learning via partial initialization.
\newblock \emph{Advances in Neural Information Processing Systems}, 34: 23309--23320.

\bibitem[{Sun et~al.(2024)Sun, Li, Li, and Ding}]{sun2024improving}
Sun, Y.; Li, Z.; Li, Y.; and Ding, B. 2024.
\newblock Improving loRA in privacy-preserving federated learning.
\newblock \emph{arXiv preprint arXiv:2403.12313}.

\bibitem[{Tao et~al.(2024)Tao, Mason, Kulkarni, and Boix}]{tao2024task}
Tao, Z.~S.; Mason, I.; Kulkarni, S.; and Boix, X. 2024.
\newblock Task arithmetic through the lens of one-shot federated learning.
\newblock \emph{arXiv preprint arXiv:2411.18607}.

\bibitem[{Touvron et~al.(2023)Touvron, Lavril, Izacard, Martinet, Lachaux, Lacroix, Rozi{\`e}re, Goyal, Hambro, Azhar et~al.}]{touvron2023llama}
Touvron, H.; Lavril, T.; Izacard, G.; Martinet, X.; Lachaux, M.-A.; Lacroix, T.; Rozi{\`e}re, B.; Goyal, N.; Hambro, E.; Azhar, F.; et~al. 2023.
\newblock Llama: Open and efficient foundation language models.
\newblock \emph{arXiv preprint arXiv:2302.13971}.

\bibitem[{Venkateswara et~al.(2017)Venkateswara, Eusebio, Chakraborty, and Panchanathan}]{venkateswara2017deep}
Venkateswara, H.; Eusebio, J.; Chakraborty, S.; and Panchanathan, S. 2017.
\newblock Deep hashing network for unsupervised domain adaptation.
\newblock In \emph{Proceedings of the IEEE conference on computer vision and pattern recognition}, 5018--5027.

\bibitem[{Wang et~al.(2024)Wang, Shen, He, Sun, Wang, Lyu, and Li}]{wang2024flora}
Wang, Z.; Shen, Z.; He, Y.; Sun, G.; Wang, H.; Lyu, L.; and Li, A. 2024.
\newblock Flora: Federated fine-tuning large language models with heterogeneous low-rank adaptations.
\newblock \emph{arXiv preprint arXiv:2409.05976}.

\bibitem[{Wei et~al.(2021)Wei, Bosma, Zhao, Guu, Yu, Lester, Du, Dai, and Le}]{wei2021finetuned}
Wei, J.; Bosma, M.; Zhao, V.~Y.; Guu, K.; Yu, A.~W.; Lester, B.; Du, N.; Dai, A.~M.; and Le, Q.~V. 2021.
\newblock Finetuned language models are zero-shot learners.
\newblock \emph{arXiv preprint arXiv:2109.01652}.

\bibitem[{Yang et~al.(2024)Yang, Long, Shen, Jiang, and Blumenstein}]{yang2024dual}
Yang, Y.; Long, G.; Shen, T.; Jiang, J.; and Blumenstein, M. 2024.
\newblock Dual-Personalizing Adapter for Federated Foundation Models.
\newblock \emph{arXiv preprint arXiv:2403.19211}.

\bibitem[{Yang et~al.(2025)Yang, Long, Zhou, Lu, Ye, and Jiang}]{yang2025federated}
Yang, Y.; Long, G.; Zhou, T.; Lu, Q.; Ye, S.; and Jiang, J. 2025.
\newblock Federated Adapter on Foundation Models: An Out-Of-Distribution Approach.
\newblock \emph{arXiv preprint arXiv:2505.01075}.

\bibitem[{Ye et~al.(2023)Ye, Xu, Wang, Xu, Chen, and Wang}]{ye2023feddisco}
Ye, R.; Xu, M.; Wang, J.; Xu, C.; Chen, S.; and Wang, Y. 2023.
\newblock Feddisco: Federated learning with discrepancy-aware collaboration.
\newblock In \emph{International Conference on Machine Learning}, 39879--39902. PMLR.

\bibitem[{Yu, Mu{\~n}oz, and Jannesari(2023)}]{yu2023federated}
Yu, S.; Mu{\~n}oz, J.~P.; and Jannesari, A. 2023.
\newblock Federated Foundation Models: Privacy-Preserving and Collaborative Learning for Large Models.
\newblock \emph{arXiv preprint arXiv:2305.11414}.

\bibitem[{Zhang et~al.(2023{\natexlab{a}})Zhang, Hua, Wang, Song, Xue, Ma, and Guan}]{zhang2023fedala}
Zhang, J.; Hua, Y.; Wang, H.; Song, T.; Xue, Z.; Ma, R.; and Guan, H. 2023{\natexlab{a}}.
\newblock Fedala: Adaptive local aggregation for personalized federated learning.
\newblock In \emph{Proceedings of the AAAI conference on artificial intelligence}, volume~37, 11237--11244.

\bibitem[{Zhang et~al.(2023{\natexlab{b}})Zhang, Vahidian, Kuo, Li, Zhang, Wang, and Chen}]{zhang2023towards}
Zhang, J.; Vahidian, S.; Kuo, M.; Li, C.; Zhang, R.; Wang, G.; and Chen, Y. 2023{\natexlab{b}}.
\newblock Towards Building the Federated GPT: Federated Instruction Tuning.
\newblock \emph{arXiv preprint arXiv:2305.05644}.

\bibitem[{Zhang et~al.(2023{\natexlab{c}})Zhang, Yang, Dai, Wang, Yu, Qu, and Xu}]{zhang2023fedpetuning}
Zhang, Z.; Yang, Y.; Dai, Y.; Wang, Q.; Yu, Y.; Qu, L.; and Xu, Z. 2023{\natexlab{c}}.
\newblock FedPETuning: When federated learning meets the parameter-efficient tuning methods of pre-trained language models.
\newblock In \emph{Annual Meeting of the Association of Computational Linguistics 2023}, 9963--9977. Association for Computational Linguistics (ACL).

\bibitem[{Zhuang, Chen, and Lyu(2023)}]{zhuang2023foundation}
Zhuang, W.; Chen, C.; and Lyu, L. 2023.
\newblock When foundation model meets federated learning: Motivations, challenges, and future directions.
\newblock \emph{arXiv preprint arXiv:2306.15546}.

\end{thebibliography}
\end{document}